\def\year{2022}\relax
\documentclass[letterpaper]{article} 
\usepackage{aaai22}  
\usepackage{times}  
\usepackage{helvet}  
\usepackage{courier}  
\usepackage[hyphens]{url}  
\usepackage{graphicx} 
\usepackage{natbib}  
\usepackage{caption} 
\DeclareCaptionStyle{ruled}{labelfont=normalfont,labelsep=colon,strut=off} 
\usepackage{algorithm}
\usepackage{algorithmic}

\usepackage{microtype}
\usepackage{graphicx}
\usepackage{subfigure}
\usepackage{booktabs} 

\usepackage{amsfonts,amssymb}       
\usepackage{nicefrac}       
\usepackage{microtype}      
\usepackage{amsmath,xcolor,colortbl,amsthm}
\usepackage{bm}
\usepackage{adjustbox}
\usepackage{enumitem}
\usepackage{multirow}

\newcommand{\thetab}{{\bm \theta}}

\newtheorem{defin}{Definition}

\definecolor{LightGreen}{rgb}{0.8,1,0.6}
\definecolor{LightBlue}{rgb}{0.5,0.9,1}
\definecolor{LightPurple}{rgb}{0.9,0.8,1}
\definecolor{Blu}{rgb}{0.7,1,1}
\definecolor{Torquoise}{rgb}{0.4,1,0.8}

\usepackage{newfloat}
\usepackage{listings}
\floatstyle{ruled}
\newfloat{listing}{tb}{lst}{}
\floatname{listing}{Listing}
\title{A Statistical Relational Approach to\\ Learning Distance-based GCNs}
\author {
    Devendra Singh Dhami\textsuperscript{\rm 1,2},
    Siwen Yan \textsuperscript{\rm 2},
    Sriraam Natarajan \textsuperscript{\rm 2}
}
\affiliations {
    \textsuperscript{\rm 1} Technical University of Darmstadt, Germany\\
    \textsuperscript{\rm 2} The University of Texas at Dallas, USA\\
    devendra.dhami@cs.tu-darmstadt.de, siwen.yan@utdallas.edu, sriraam.natarajan@utdallas.edu
}

\usepackage{bibentry}

\begin{document}

\maketitle

\begin{abstract}
We consider the problem of learning distance-based Graph Convolutional Networks (GCNs) for relational data. Specifically, we first embed the original graph into the Euclidean space $\mathbb{R}^m$ using a {\em relational density estimation technique} thereby constructing a secondary Euclidean graph. The graph vertices correspond to the target triples and edges denote the Euclidean distances between the target triples. We emphasize the importance of learning the secondary Euclidean graph and the advantages of employing a distance matrix over the typically used adjacency matrix. Our comprehensive empirical evaluation demonstrates the superiority of our approach over $\mathbf{12}$ different GCN models, relational embedding techniques and rule learning techniques.
\end{abstract}

\section{Introduction}
Statistical Relational Learning (SRL) \cite{getoor2007introduction,raedt2016statistical} combines the power of probabilistic models to handle uncertainty with the ability of relational models to faithfully capture the rich domain structure. One of the key successes of these models lie in the task of knowledge base population, specifically, link prediction and node classification. While successful, most methods make several simplifying assumptions -- presence of supervision in the form of labels, closed-world assumption, presence of only binary relations and most importantly, in many cases, the presence of hand-crafted domain rules.

We go beyond these assumptions and inspired by the recent success of Graph Convolutional Networks (GCNs)~\cite{defferrard2016convolutional, kipf2017semi}, develop a {\em new framework for relational GCNs}. This framework has two key steps: (1) create a secondary Euclidean graph from the original graph by {\em learning} rules from one-class data, i.e., from the positive and negative annotations of the target relation separately. The next step is to {\em convert} these rules into observed features i.e., instantiates and counts the number of times the rules fire and computes the distance matrix, and (2) finally, it {\em trains} a GCN using the observed features and the distance matrix. For the first step, our method employs a one-class density estimation method that employs a tree-based distance metric to learn relational rules iteratively. Hence, we call the framework as {\em Relational Density Distance-based GCN} (RD$^2$GCN). Since the two different steps of learning the relational rules and training the  GCN employ the same set of positive examples, a richer representation of the combination of the attributes, entities and their relations is obtained. While previous methods used the features as the observed layer, RD$^2$GCN uses the rules as the observed layer. This has the added advantage of the latent layer being richer -- it combines the instantiations of first-order rules themselves allowing for a richer representation. We hypothesize and show empirically that this is specifically useful when employed on link prediction tasks. Although work exists on generating similarity graphs using GNNs~\cite{bai2018convolutional,bai2019simgnn,li2019graph}, ours is the first method to use GCNs on induced similarities graphs allowing for use of richer features.

We make a few key contributions: (1) We develop the first relational GCN capable of utilizing the different densities of the data separately. (2) Going beyond using carefully designed hand-crafted rules, our method learns rules automatically to construct a secondary graph and  constructs the GCN. These two steps are conditioned on the required task and allow for a better classifier and thus can learn with {\em smaller data}. (3) RD$^2$GCN can handle arbitrary relations -- not simple binary relations that most methods use. Given that our base learner employs a logic learner, the relations can be $n-$ary. (4) We show the advantages of using distance matrices and Euclidean distance to construct the distance matrix.
Our evaluation across {\bf $\mathbf{12}$ different baselines and $3$ different data sets clearly demonstrates the effectiveness of RD$^2$GCN}.

\section{Background and Related work}
\paragraph{Notations:} A (logical) \textbf{predicate} is of the form $\mathcal{R}(t_1, \dots, t_k)$ where $\mathcal{R}$ is a relation and the arguments $t_i$ are {\bf entities}. A \textbf{substitution} is of the form $\thetab = \{\langle l_1, \dots, l_k \rangle/ \langle t_1, \dots, t_k\rangle\}$ where $l_i$s are logical variables and $t_i$s are terms. A \textbf{grounding} of a predicate with variables $l_1, \dots, l_k$ is a substitution $\{\langle l_1, \dots, l_k \rangle \slash \langle L_1, \dots, L_k \rangle\}$\footnote{We use uppercase for relations/groundings and lowercase for variables.} mapping each of its variables to a constant in the domain of that variable. A knowledge base $\mathcal{B}$ consists of (1) entities: a finite domain of objects $\mathcal{O}$, (2) relations: a set of predicates describing the attributes and relationships between objects $\in \mathcal{O}$, and (3) an interpretation assigning a truth value to every grounding. 

\paragraph{Relational Density Estimation:} A common issue in many real-world relational knowledge bases is that only true instances of any relation(s) are labeled while the false instances are not explicitly identified. Consequently \textit{\textbf{closed-world assumption}} is applied to sample negative instances. While reasonable, this is a strong assumption particularly when the number of positively labeled examples $\ll$ negatively labeled examples. In the relational one-class classification \cite{khot2014relational} method, given a set of labeled examples, a distance measure is used to perform one-class classification, which involves two levels of combinations: tree-level due to learning multiple trees and instance-level due to the predicates containing variables and different instances for each target. For example, in learning {\em advisedBy(S,P)} the first tree could consider the courses and the second could consider the publications. The tree-level combining function combines the results from these two trees. Now the student could potentially publish several papers, or register in multiple courses and inside each tree, these different instances are combined using the instance level combining function~\cite{jaeger2007RBNs,natarajan2008Combining}.
In the tree-level distance computation, the distance between the current unlabeled example $u$ is calculated from a labeled example in all the learned first-order trees. Now the final distance is simply the weighted combination of the individual tree-level distances: $D(l_1, \, u)=\sum_i \, \beta_i \, d_i(l_1, u)$ where $\beta_i$ is the weight of the $i^{th}$ tree and $\sum_i \beta_i=1, \beta_i \geq 0$.  These tree distances are then combined to get an overall distance between the current example and all the labeled examples $l_j$, $E(u \not \in \texttt{class}) \, = \, \sum_j\, \alpha_j D(l_j, \,u)$, where $\alpha_j$ is the weight of the labeled example $l_j$ and  $\sum \alpha_j=1, \alpha_j \geq 0$. 

\paragraph{Knowledge Graph Embeddings (KGEs):} Recently, several successful methods for learning embeddings of large knowledge bases have been developed~\cite{wang2017knowledge, cai2018comprehensive}. 
Several of these approaches such as TransE \cite{bordes2013translating}, TransH \cite{wang2014knowledge}, TransG \cite{xiao2016transg} and KG2E \cite{he2015learning}, to name a few, can be grouped into translational distance models that focus on minimizing a distance based function under some constraints or using regularizing factors between entities and relations. More recent approaches extend these translation approaches by embedding the knowledge graphs into more complex spaces such as the hyperbolic space \cite{balazevic2019multi,kolyvakis2019hyperkg} and the hypercomplex space \cite{zhang2019quaternion,sun2019rotate}.  Another important class of approaches such as RESCAL \cite{nickel2011three}, DistMult \cite{yang2015embedding}, TuckER \cite{balavzevic2019tucker}, HypER \cite{balavzevic2019hypernetwork} and HolE \cite{nickel2016holographic} focus on various compositional operators for the entities and relations in the knowledge graph. 
\paragraph{Graph Convolutional Networks (GCNs):} 
Graph Convolutional Networks (GCNs) \cite{kipf2017semi} generalize convolutional neural network models to graph-structured data sets where each convolution layer in the GCN applies a graph convolution i.e. a spectral filtering of the graph signal (the feature matrix of the graph) via the Graph Fourier Transform.
The main reason behind the success of GCNs is that they exploit two key types of information: node feature descriptions ($x_i$) and node neighborhood structure (captured through the adjacency matrix $\mathcal{A}$ of the graph). 
While successful, GCNs cannot directly be applied on multi-relational data/networks and require propositionalization techniques. Consequently, relational GCNs~\cite{schlichtkrull2018modeling} construct a latent representation of the entities explicitly and a tensor factorization then exploits these representations for the prediction tasks. We take an alternative approach based on a successful SRL approach ~\cite{khot2014relational,lao2010relational} to develop novel combinations of the entities and their relationships to construct richer latent representations. As we demonstrate empirically, this leads to superior predictive performance. In addition, the use of {\em relational rules as the observed layer of the GCN makes them more interpretable/explainable than the tensor factorization approach}.

\begin{figure}[!t]
    \centering
    \includegraphics[width =1.0\columnwidth]{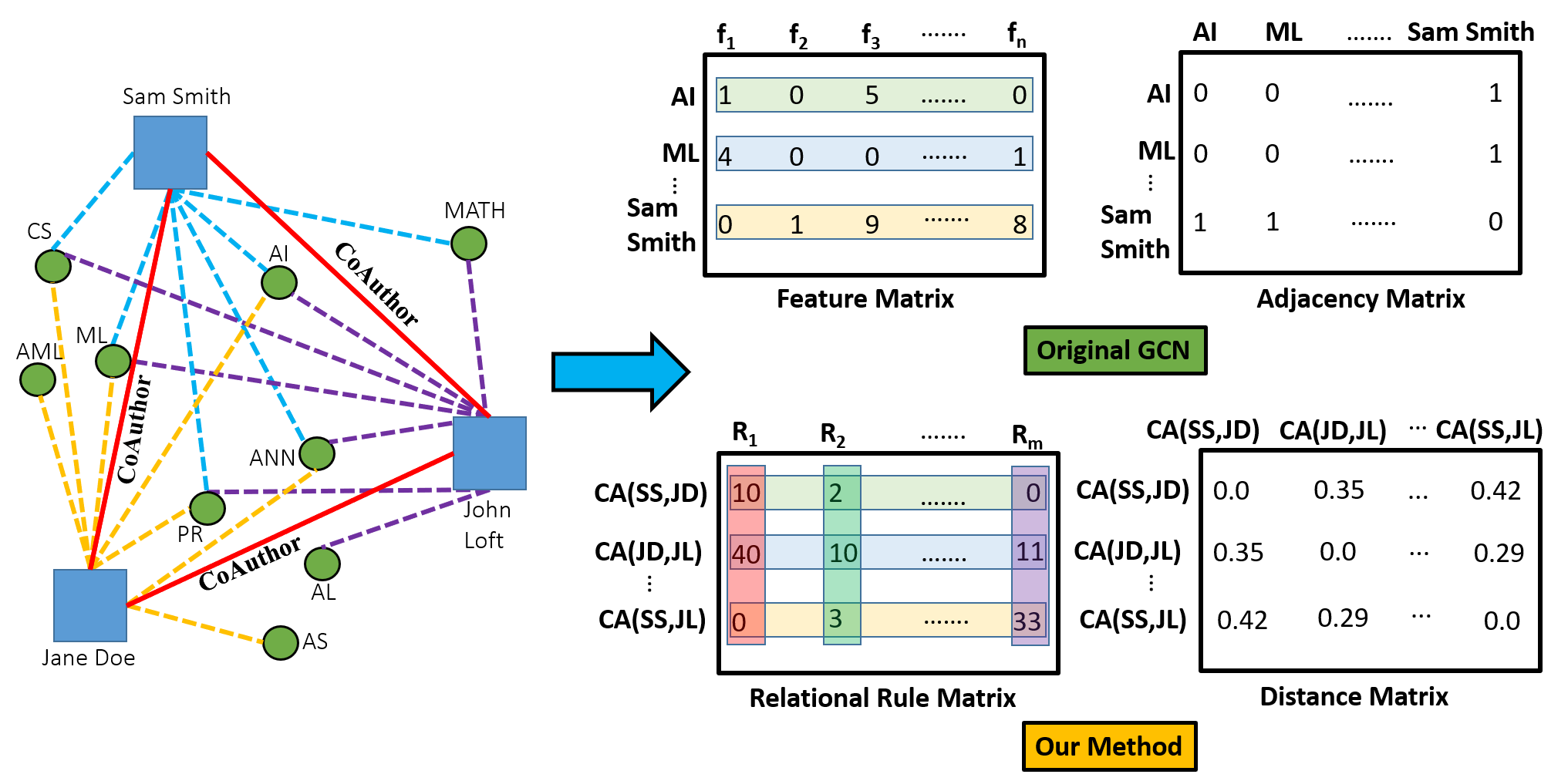}
    \caption{Difference between GCN \cite{kipf2017semi} and our method with an example input graph. Here, CA is the CoAuthor relation to be predicted. The relational rule matrix is obtained by counting the number of satisfied groundings of the obtained first-order rules (R$_1$ - R$_m$) wrt the query variables and is a richer representation of the graph structure.}
    \label{fig:method}
\end{figure}

\section{Relational Density Distance-based GCNs}
\label{sec:rgcn}
Direct application of GCNs cannot fully exploit the inherent structures inside a multi-relational graph. Consequently, they need significant engineering to construct the propositionalized features. Motivated by this, we propose a principled extension to the GCN that models large multi-relational networks faithfully. While a recent work R-GCN~\cite{schlichtkrull2018modeling} extends GCNs to relational domains, it is still limited to graphs represented as {\em (subject; predicate; object)} triples and requires multiple adjacency matrices for handling multi-relational data. We propose a novel and a more general approach that is not limited by assumptions about the multi-relationality of the data and can handle general multi-graphs and hypergraphs without loss of information. We can now formally define our model and its components.

\begin{defin}[\textbf{Secondary Euclidean Graph}]
\label{def:eucgraph}
A secondary Euclidean graph consists of a set of vertices and edges where the vertices correspond to the query variable in the relational data set and the edges constitute the Euclidean distance between each pair of vertices.
\end{defin}

\begin{defin}[\textbf{RD$^2$GCN}]
\label{def:ROCGCN}
Given a knowledge base/relational graph $\mathcal{B}$ and a function $\phi: \mathcal{B} \mapsto \mathbb{R}^m$, such that $\phi$($\mathbb{B}$) = $\mathfrak{E} \in \mathbb{R}^m$, RD$^2$GCN $\mathfrak{G}$ is a graph convolutional network defined over $\mathfrak{E}$ and $Euc(\mathfrak{E})$ i.e. the secondary Euclidean graph.
\end{defin}

\begin{defin}[\textbf{Relational Rule Matrix}]
\label{def:rrm}
A relational rule matrix $\mathcal{X}$ contains the node feature descriptors $x_i \in \mathfrak{E}$ for a Euclidean graph.
\end{defin}

\begin{defin}[\textbf{Distance Matrix}]
\label{def:dm}
A distance matrix $\mathcal{D}$ contains the euclidean distances between the node feature descriptors $\in \mathcal{X}$ such that $Euc(\mathfrak{E}) \in \mathcal{D}$.
\end{defin}

\begin{figure}[!t]
    \centering
    \includegraphics[width =1.0\columnwidth]{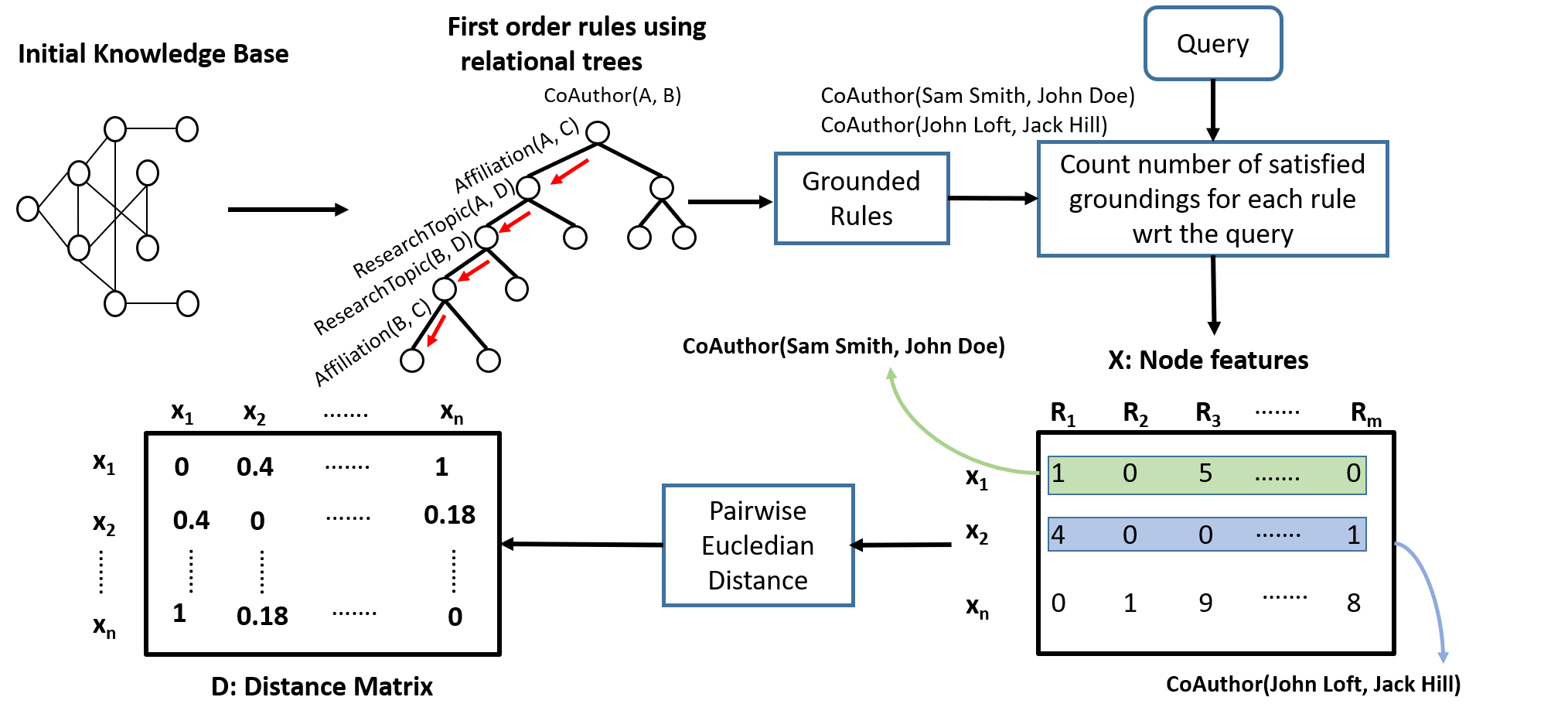}
    \caption{Relational Rule Matrix $\mathcal{X}$ and Distance Matrix $\mathcal{D}$ construction for RD$^2$GCN. First-order rules are learned from the given knowledge bases which are then grounded and satisfied groundings are counted to form $\mathcal{X}$. }
    \label{fig:rgcn1}
\end{figure}

Given a knowledge base $\mathcal{B}$, we first learn a set of first-order rules that captures the relations between the domain predicates. The intuition is that these first-order rules can be viewed as {\em higher-order features} that connect entities and their attributes. Particularly, when learned for a specific classification task, these features can be both predictive and informative. Given that they are typically conjunctions of relational features (attributes of entities and relationships), they have the added advantage of being interpretable. Our hypothesis, that we verify empirically is that these rules can potentially yield richer latent representations than a relational GCN that simply uses the entity and relationship information.
\noindent \fbox{
 \parbox{0.96\columnwidth}{
\emph{Our key contribution is a two-step process of constructing the link prediction problem as a prediction problem in a secondary Euclidean graph where vertices correspond to target triple rather than individual entities. We learn a relational rule matrix and then build a distance matrix to use for GCN-computations (see Figure \ref{fig:method}).}\\
 }}
 
\subsection{Embedding Original Graph to $\mathbb{R}^m$: Creating a Euclidean Graph}
We now outline the required steps to embed the original graph to a Euclidean space $\mathbb{R}^n$ thereby creating a secondary Euclidean graph. The nodes of the Euclidean graph consists of the target triple with the node features forming the relational rule matrix $\mathcal{X}$ and the edges connecting the nodes are the Euclidean distances thus forming the distance matrix $\mathcal{D}$. It is clear from def. \ref{def:eucgraph}-\ref{def:dm} that we just need $\mathcal{X}$ and $\mathcal{D}$ to represent a secondary Euclidean graph. Figure \ref{fig:rgcn1} shows the construction of $\mathcal{X}$ and $\mathcal{D}$.

\begin{figure}[t]
    \centering
    \includegraphics[width =1.0\columnwidth]{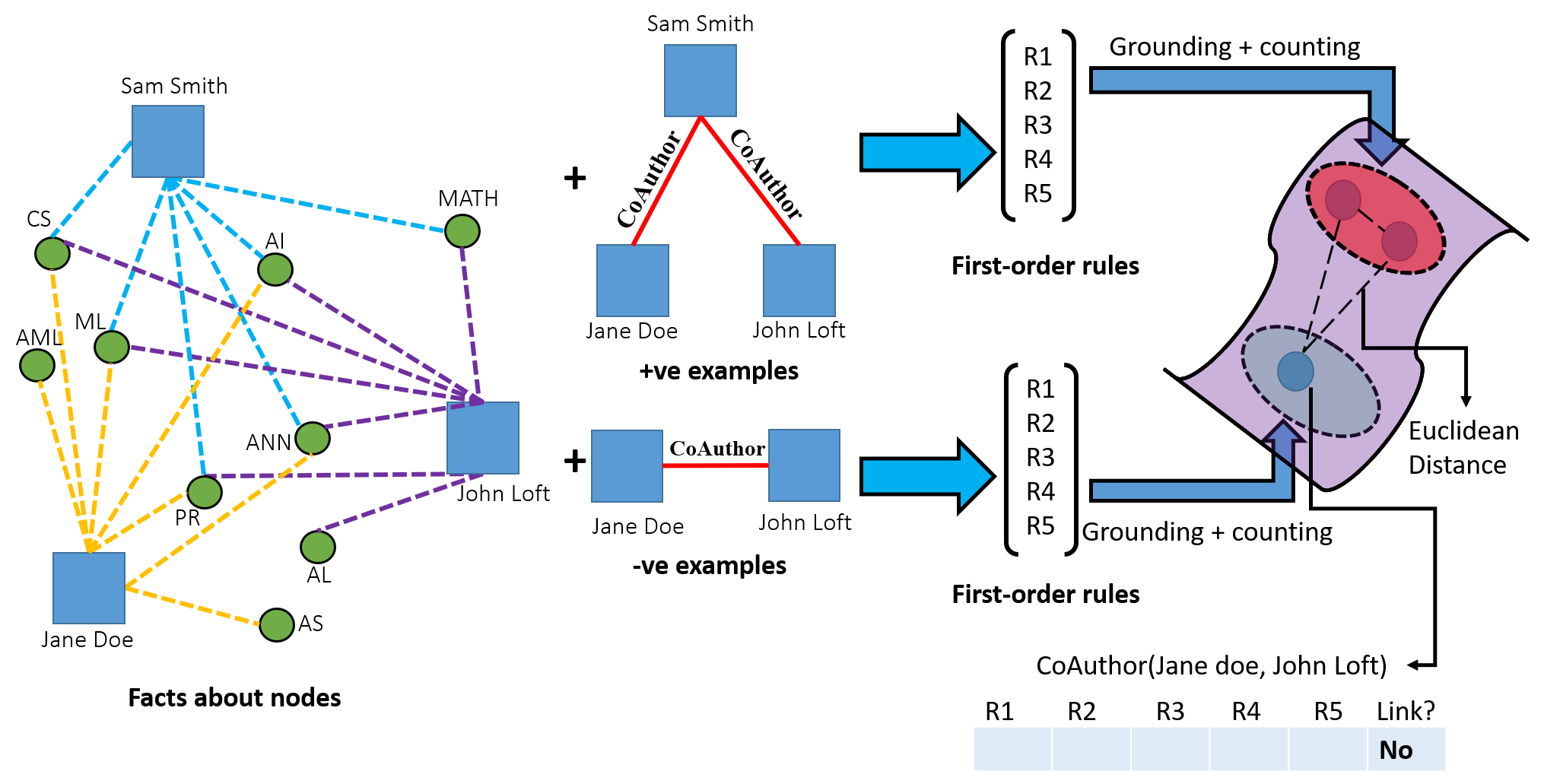}
    \caption{Learning secondary Euclidean graph (nodes) for ICML data set. Learning the +ve and -ve rules and thus features separately result in more discriminative secondary graph nodes with the +ve nodes closer to each other and distant from the -ve node.}
    \label{fig:mani}
\end{figure}

\textbf{Step 1: Rule Learning using Density Estimation}:  Inspired by the success of learning only from positive examples in relational domains~\cite{khot2014relational}, we learn first-order rules using relational density estimation (which forms $\phi$ in def \ref{def:ROCGCN}) and learn from {\bf both the positive and negative examples separately}. The intuition behind using a density estimation method is that \emph{learning first-order rules for positive and sampled negative examples independently can result in better utilization of the search space thereby (potentially) learning more discriminative features}. Figure \ref{fig:mani} shows an example of learning such discriminative features for a ``Co-Author" data set. The density estimation approach uses a tree-based distance measure that iteratively introduces newer features (as short rules) that covers more positive examples.

Thus, we construct a relational graph manifold, by treating relational examples as nodes and connect ones that are close or similar to each other in the neighborhood. The similarity can be measured by learning a tree-based distance between relational examples and is inversely proportional to the depth $d$ of least common ancestor (LCA) of the pair of examples, say $l_1,l_2$ being considered,
\begin{equation}
    d(l_1, l_2)=  
    \begin{cases}
         0,  & \mathsf{LCA}(l_1, \, l_2) \,\, \textrm{is leaf}; \\
        e^{-\lambda \cdot \mathsf{depth}(\mathsf{LCA}(l_1, l_2))}, & \text{otherwise},
\end{cases}
\label{eq:lca}
\end{equation}
where $\lambda > 0$ ensures that distance decreases (i.e., similarity increases) as the depth increases.

We learn a tree-based distance iteratively to introduce new relational features that perform one-class classification. The left-most path in each relational tree is a conjunction of predicates, that is, a clause, which can be used as a relational feature. The splitting criteria is the squared error over the examples and the goal is to minimize squared error in each node.
We consider only the left-most path of every tree constructed due to the fact that the right paths all involve negations. Thus a single rule is extracted from each tree. The final set of ``trees" is simply a list of ``first-order rules" and thus are interpretable and explainable. We now present some example first order rules learned by density estimation for 2 of our considered data sets. The first two rules for each data set are learnt for the positive examples and the next two are learnt for negative examples.
\paragraph{\textit{Data set: Drug-Drug Interactions}}
\begin{enumerate}
    \item \underline{Rule}: Interacts($d_1$, $d_2$) $\implies$ TransporterSubstrate($d_1$, $tr_1$) $\wedge$ TransporterSubstrate($d_2$, $tr_1$) $\wedge$ EnzymeInhibitor($d_1$, $e_1$) $\wedge$ EnzymeInhibitor($d_2$, $e_1$)\\
    \item \underline{Rule}: Interacts($d_1$, $d_2$) $\implies$ EnzymeInducer($d_1$, $e_1$) $\wedge$ EnzymeSubstrate($d_2$, $e_1$) $\wedge$ EnzymeInducer($d_2$, $e_2$) $\wedge$ EnzymeInducer($d_1$, $e_2$)\\
    \item \underline{Rule}: Interacts($d_1$, $d_2$) $\implies$ TargetInhibitor($d_1$, $t_1$) $\wedge$ TargetInhibitor($d_2$, $t_2$) $\wedge$ TransporterSubstrate($d_1$, $tr_1$)\\
    \item \underline{Rule}: Interacts($d_1$, $d_2$) $\implies$ TargetAgonist($d_1$, $t_1$) $\wedge$ TargetAgonist($d_2$, $t_2$) $\wedge$ Transpor\\terInducer($d_1$, $tr_1$) $\wedge$ TransporterInducer($d_2$, $tr_2$)\\
\end{enumerate}

\paragraph{\textit{Data set: ICML CoAuthor}}
\begin{enumerate}
    \item \underline{Rule}: CoAuthor($p_1$, $p_2$) $\implies$ Affiliation($p_1$, $a_1$) $\wedge$ Affiliation($p_2$, $a_1$) $\wedge$ ResearchTopic($p_1$, $topic_1$) $\wedge$ ResearchTopic($p_2$, $topic_1$)\\
    \item \underline{Rule}: CoAuthor($p_1$, $p_2$) $\implies$ ResearchTopic($p_2$, $``Mathematical\_Optimization"$) $\wedge$ ResearchTopic($p_1$, $``Pattern\_Recognition"$) $\wedge$ ResearchTopic($p_1$,$topic_1$) $\wedge$ \\ResearchTopic($p_2$,$topic_1$)\\
    \item \underline{Rule}: CoAuthor($p_1$, $p_2$) $\implies$ ResearchTopic($p_1$, $``Pattern\_Recognition"$) $\wedge$ Research\\Topic($p_2$, $``Mathematical\_Optimization"$)\\
    \item \underline{Rule}: CoAuthor($p_1$, $p_2$) $\implies$ Affiliation($p_1$, $``University\_of\_California\_Berkeley"$) $\wedge$ Affiliation($p_2$, $``Simons\_Institute"$)\\
\end{enumerate}

\textbf{Step 2: Relational Rule Matrix and Distance Matrix Construction:} The learned first-order rules are then grounded to obtain all the instantiations of these rules. The counts of each feature, i.e., the count of the number of times a target example (the coauthor relation between the target entities) is satisfied in every first-order rule is obtained which forms our relational rule matrix $\mathcal{X}$. In spirit, this is similar to MLNs~\cite{richardson2006markov} that counts the instances to obtain a marginal distribution. Instead of using the counts to compute marginals, we use them in the matrices. For example, the learned first-order rule from true instances
\[
\scalebox{1.0}{$
\begin{array}{cl}
& \mathtt{CoAuthor(person_1, person_2)} \, \Leftarrow \, \\
& \mathtt{Affiliation(person_1, university_1)}\\
& \wedge \mathtt{Affiliation(person_2, university_1)} \\
& \wedge \mathtt{ResearchTopic(person_1, topic_1)}\\
& \wedge \mathtt{ResearchTopic(person_2, topic_1)}. 
\end{array}$}
\]
implies that if two persons have the same affiliation and their research interests lie in same topics, then they are likely to coauthor. Suppose the given target entities are $person_1$ = ``Jane Doe" (JD) and $person_2$ = ``Sam Smith"(SS). The partially grounded first-order rule can then be written as
\[
\scalebox{1.0}{$
\begin{array}{cl}
& \mathtt{CoAuthor(JD, SS)} \, \Leftarrow \, \\
& \mathtt{Affiliation(JD, university_1)}\\
& \wedge \mathtt{Affiliation(SS, university_1)} \\
& \wedge \mathtt{ResearchTopic(JD, topic_1)}\\
& \wedge \mathtt{ResearchTopic(SS, topic_1)}. 
\end{array}$}
\]
Then substitutions for all the other entities within the first-order rule are performed and checked whether the substituted first-order rule is satisfied in the groundings. For example, the substitution  $\thetab = \{\langle university_1, topic_1\rangle/ \langle UCB, Artificial\:Intelligence\rangle\}$
is satisfied but the substitution $\thetab = \{\langle university_1, topic_1\rangle/ \langle UCB, Computer\:Networks\rangle\}$ is not satisfied. Since there can be multiple values taken by $topic_1$ that can satisfy the first-order rule, the count of all such satisfied groundings becomes a feature value for the target query {\tt CoAuthor(Jane\:Doe, Sam Smith)}. Thus using this satisfiability count  we obtain a feature set $\mathcal{X}$ of size $n \times k$ where n is number of target queries and k is number of first-order rules that represent the node features.\\
In order to obtain the distance matrix $\mathcal{D}$ a pairwise euclidean distance of all the node feature descriptors i.e. the counts $x_i \in \mathcal{X}$ is computed. Thus, every element $d_{ij} \in \mathcal{D}$,
\begin{equation}
d_{ij}  = \|x_i - x_j\|=\sqrt{\|x_i\|^2+\|x_j\|^2-2 \cdot x_i \cdot x_j}
\end{equation}

\subsection{Euclidean Graph GCN}
The original GCN formulation \cite{kipf2017semi} requires an adjacency matrix $\mathcal{A}$ to perform the layer-wise propagation. Instead of building the adjacency matrix from the relation triples, we use the computed geometric distance matrix $\mathcal{D}$, which is a richer structure~\cite{rouvray1979chemical,cvetkovic1980spectra}, and use it as an approximation to the adjacency matrix for the GCN. To obtain this approximation, we perform the following steps:

\textbf{[1]:} A threshold, t, is set as the average of all the distances (since the distance matrix is symmetric, the average is calculated from the upper-right part). 

\textbf{[2]:} $\forall d_{ij} \in \mathcal{D}$, new distances are computed as $\hat{d}_{ij}= d_{ij}/t$ and $\hat{d}_{ij}>$ 1 is set as 1: a far-away case.

\textbf{[3]:} Since the higher values in $\mathcal{D}$ represent nodes that are far as opposed to the $\mathcal{A}$ where the higher values i.e. 1 represents the nodes adjacent to each other, the distance between nodes is subtracted from 1 i.e. $\hat{d}_{ij}= 1 - \hat{d}_{ij}$. This is similar to $\mathcal{A}$ with $\hat{d}_{ij}$= 1 representing that two nodes are connected and $\hat{d}_{ij}$= 0 representing that two nodes are not connected with the only difference being the presence of values 0 $< \hat{d}_{ij} <$ 1 that denote the closeness of two nodes.

For a GCN with M layers, the layer wise propagation rule for the layer $l$ $\in$ M can now be written as,
\begin{equation}
    f(H^{(l)},\mathcal{D}) = \sigma(\mathcal{D}H^{(l)}W^{(l)})
\end{equation}
where $H^{(0)}$ is the input layer i.e. the relational rule matrix $\mathcal{X}$ with $H^{(1)} \dots H^{(M-1)}$ being the hidden layers. Since we replace $\mathcal{A}$ with $\mathcal{D}$ before the symmetric normalization and addition of self loops, these operations are now performed on $\mathcal{D}$.  The updated propagation rule is,
\begin{equation}
    f(H^{(l)},\mathcal{D}) = \sigma(\hat{\mathcal{N}}^{\frac{-1}{2}} \hat{\mathcal{D}} \hat{\mathcal{N}}^{\frac{-1}{2}} H^{(l)}W^{(l)})
\end{equation}
such that $\hat{\mathcal{D}}=\mathcal{D}$ + $\mathcal{I}$ where $\mathcal{I}$ is the identity matrix and $\hat{\mathcal{N}} \in \mathbb{R}$ is the diagonal weighted node degree matrix of $\hat{\mathcal{D}}$.
In summary, we learn first-order rules from separate densities independently, in the process constructing a secondary graph consisting of query variable as nodes. These learned rules are then grounded resulting in richer representation than simple node features. For obtaining distance between target triples to define adjacency, we use pairwise Euclidean distance. We present rigorous empirical evaluations next.

\begin{table*}[!h]
    \centering
    \caption{Properties of data sets. 
    }
    \begin{tabular}{|c|c|c|c|c|c|c|}
        \hline
         Task & Data Set & \# Relations & \# Facts & \#+ve Examples & \#-ve Examples & \# Rules\\
        \hline
        \multirow{3}{*}{Link Prediction} & \emph{ICML'18} & 4 & 1395 & 155 & 6498 & 7\\
        &\emph{ICLR} & 4 & 4730 & 990 & 10000 & 7\\
        &\emph{DDI} & 14 & 1774 & 2832 & 3188 & 25\\
        \hline
        \end{tabular}
    \label{tab:prop}
 \end{table*}
 
 \section{Experimental Results}
We consider 3 relational data sets for link prediction(Table~\ref{tab:prop}). \emph{ICML'18} consists of papers from ICML 2018, \emph{ICLR} consists of papers from ICLR (2013-2019) and the prediction task is whether two people are coauthors for both data sets. Both of these data sets are extracted from the Microsoft Academic Graph (MAG) \cite{sinha2015overview}. \emph{DDI} is a drug-drug interaction data set \cite{dhami2018drug} and the goal is to predict whether two drugs interact. A limitation of our work is that we cannot handle multiple query variables without joint learning where one could consider every relation as the query variable in different rule learning steps to obtain embeddings w.r.t all relations and use them for the knowledge base completion tasks. We leave it as a future work. 

We first learn first-order logic rules using relational density estimation~\cite{khot2014relational} from positive examples. The number of rules learned each for positive and negative examples is shown in table \ref{tab:prop}.The relational rule matrix $\mathcal{X}$ and the distance matrix $\mathcal{D}$ are then obtained. 
We aim to answer the following questions: 
\begin{enumerate}
    \item[] \textbf{Q1:} How does our method perform on data sets that have few examples?
    \item[] \textbf{Q2:} Is learning a secondary graph structure useful?
    \item[] \textbf{Q3:} Can the combination of SRL with deep models such as GCN result in better predictive models?
    \item[]  \textbf{Q4:} How does rule learning from relational density estimation compare with other rule learning methods?
    \item[] \textbf{Q5:} What is the effect of different distance measures on the RD$^2$GCN performance?
    \item[] \textbf{Q6:} How sensitive is RD$^2$GCN to the choice of parameters?
\end{enumerate}    
\subsection{Baselines}
\textbf{Link Prediction:} We compare RD$^2$GCN, to {\bf 12 embedding baselines} in 3 categories. 

\underline{\textit{1. Rule learning methods}}: \textbf{Gaifman models} \cite{niepert2016discriminative}: uses Gaifman locality principle \cite{gaifman1982local} to enumerate all \emph{hand-written first-order rules} within the neighborhood of the target/query variables. After obtaining the counts for the satisfied grounded handwritten rules logistic regression is used for prediction. \textbf{Neural-LP} \cite{yang2017differentiable}: learns first-order rules by extending the probabilistic differentiable logic system TensorLog \cite{cohen2016tensorlog}. \textbf{metapath2vec} \cite{dong2017metapath2vec}: generates random walks with user defined meta paths and uses a heterogeneous skip-gram model to generate embeddings. \textbf{PRAGCN:} makes use of relational random walks (PRA) \cite{lao2010relational} to learn the first-order rules~\cite{kaur2019neural} and obtain the features as described in our method. The learned features are then passed on to a GCN. \textbf{Node+LinkFeat} \cite{toutanova2015observed} (N+LF): is obtained by running logistic regression over the learned propositional features.

\underline{\textit{2. Relational embedding methods}}: \textbf{ComplEx} \cite{trouillon2016complex}: proposes a latent factorization approach in multi-relational graphs. We use the ComplEx implementation in the AmpliGraph python library\footnote{https://github.com/Accenture/AmpliGraph}. \textbf{ConvE} \cite{dettmers2018convolutional}: uses convolutions over embeddings and fully connected layers to model interactions between input entities and relationships. We use ConvE from AmpliGraph python library. \textbf{SimplE} \cite{kazemi2018simple}: adapts the concept of Canonical Polyadic decomposition and learns two dependent embeddings for each entity and relation to obtain a similarity score for each triple to perform link prediction. We use the tensorflow implementation\footnote{https://github.com/Mehran-k/SimplE}.

\underline{\textit{3. GCN based methods}}: \textbf{Relational GCN} \cite{schlichtkrull2018modeling}: extends GCN to the relational setting. and can handle different weighted edge types i.e. relations. It uses a 2 step message passing technique to learn new node representations which are then fed to a factorization method, DistMult \cite{yang2015embedding}. We use the tenserflow implementation\footnote{https://github.com/MichSchli/RelationPrediction}. \textbf{CompGCN} \cite{vashishth2020composition}: jointly embeds both nodes and relations in a graph and we use PyTorch implementation\footnote{https://github.com/malllabiisc/CompGCN}.

\subsection{Results}
For RD$^2$GCN, we use a GCN with 2 hidden layers each with dimension = 16 with a drop out layer between the 2 graph convolutional layers. To introduce non-linearity, we use {\em relu} between input and hidden layers and to score queries, we use $log\_softmax$ function. The examples for training, validation and testing are randomly sampled without replacement. For neural embedding baselines, since they are trained on true relations, the positive examples are randomly split to $\langle 60\%, 10\%, 30\% \rangle$ in training, validation and testing respectively. To obtain the different metrics for the neural embedding baseline, the scores for each pair of nodes in the test examples were thresholded by the average of the obtained scores. If the score between pair of nodes $\geq$ average score the link is predicted to be true. We run our experiments on a GPU with 8 GeForce GTX 1080 Ti cards.

\begin{table}
\begin{minipage}[t]{\columnwidth}
    \centering
    \caption{Results for Link prediction.}
    \scalebox{0.7}{
    \begin{tabular}{|c|c|c|c|c|c|c|}
        \hline
         Data & Methods & Recall & Precision & F1 & AUC-PR\\
        \hline
        \multirow{11}{*}{\emph{ICML'18}} & Gaifman & \cellcolor{LightBlue}0.10 & \cellcolor{LightBlue}0.16 & \cellcolor{LightBlue}0.174 & \cellcolor{LightBlue}0.127\\
        & Neural-LP$_3$ & \cellcolor{LightBlue}\textbf{0.927} & \cellcolor{LightBlue}0.024 & \cellcolor{LightBlue}0.047 & \cellcolor{LightBlue}0.267\\
        & Neural-LP$_{10}$ & \cellcolor{LightBlue}0.891 & \cellcolor{LightBlue}0.035 & \cellcolor{LightBlue}0.069 & \cellcolor{LightBlue}0.143\\
        & metapath2vec & \cellcolor{LightBlue}0.836 & \cellcolor{LightBlue}0.209 & \cellcolor{LightBlue}0.335 & \cellcolor{LightBlue}0.286 \\
        & PRAGCN & \cellcolor{LightBlue}0.0 & \cellcolor{LightBlue}0.0 & \cellcolor{LightBlue}0.0 & \cellcolor{LightBlue}0.512\\
        \cline{2-6}
        & ComplEx & \cellcolor{LightGreen}0.85 & \cellcolor{LightGreen}0.013 & \cellcolor{LightGreen}0.03 & \cellcolor{LightGreen}0.04\\
        & ConvE & \cellcolor{LightGreen}0.636 & \cellcolor{LightGreen}0.01 & \cellcolor{LightGreen}0.02 & \cellcolor{LightGreen}0.015\\
        & SimplE & \cellcolor{LightGreen}\textbf{0.927} & \cellcolor{LightGreen}0.012 & \cellcolor{LightGreen}0.023 & \cellcolor{LightGreen}0.128\\
        & N+LF & \cellcolor{LightGreen}0.379 & \cellcolor{LightGreen}\textbf{1.0} & \cellcolor{LightGreen}0.549 & \cellcolor{LightGreen}0.396\\
        \cline{2-6}
        & R-GCN & \cellcolor{LightPurple}0.636 & \cellcolor{LightPurple}0.07 & \cellcolor{LightPurple}0.13 & \cellcolor{LightPurple}0.13\\
        & CompGCN & \cellcolor{LightPurple}0.727 & \cellcolor{LightPurple}0.022 & \cellcolor{LightPurple}0.044 & \cellcolor{LightPurple}0.185\\
        \cline{2-6}
        & \textbf{RD$^2$GCN} & \cellcolor{Blu}0.389 & \cellcolor{Blu}\textbf{1.0} & \cellcolor{Blu}\textbf{0.561} & \cellcolor{Blu}\textbf{0.556} \\
        \hline
        \multirow{11}{*}{\emph{ICLR}} & Gaifman & \cellcolor{LightBlue}0.564 & \cellcolor{LightBlue}0.795 & \cellcolor{LightBlue}0.66 & \cellcolor{LightBlue}0.488\\
        & Neural-LP$_3$ & \cellcolor{LightBlue}0.939 & \cellcolor{LightBlue}0.308 & \cellcolor{LightBlue}0.463 & \cellcolor{LightBlue}0.421\\
        & Neural-LP$_{10}$ & \cellcolor{LightBlue}\textbf{0.987} & \cellcolor{LightBlue}0.275 & \cellcolor{LightBlue}0.429 & \cellcolor{LightBlue}0.453\\
        & metapath2vec & \cellcolor{LightBlue}0.828 & \cellcolor{LightBlue}0.338 & \cellcolor{LightBlue}0.480 & \cellcolor{LightBlue}0.641\\
        & PRAGCN & \cellcolor{LightBlue}0.0 & \cellcolor{LightBlue}0.0 & \cellcolor{LightBlue}0.0 & \cellcolor{LightBlue}0.544\\
        \cline{2-6}
        & ComplEx & \cellcolor{LightGreen}0.269 & \cellcolor{LightGreen}0.032 & \cellcolor{LightGreen}0.057 & \cellcolor{LightGreen}0.105\\
        & ConvE & \cellcolor{LightGreen}0.677 & \cellcolor{LightGreen}0.037 & \cellcolor{LightGreen}0.069 & \cellcolor{LightGreen}0.054\\
        & SimplE & \cellcolor{LightGreen}0.973 & \cellcolor{LightGreen}0.054 & \cellcolor{LightGreen}0.102 & \cellcolor{LightGreen}0.535\\
        & N+LF & \cellcolor{LightGreen}0.97 & \cellcolor{LightGreen}\textbf{1.0} & \cellcolor{LightGreen}\textbf{0.984} & \cellcolor{LightGreen}\textbf{0.972}\\
        \cline{2-6}
        & R-GCN & \cellcolor{LightPurple}0.667 & \cellcolor{LightPurple}0.783 & \cellcolor{LightPurple}0.720 & \cellcolor{LightPurple}0.763\\
        & CompGCN & \cellcolor{LightPurple}0.906 & \cellcolor{LightPurple}0.719 & \cellcolor{LightPurple}0.802 & \cellcolor{LightPurple}0.912\\
        \cline{2-6}
        & \textbf{RD$^2$GCN}& \cellcolor{Blu}0.594 &  \cellcolor{Blu}\textbf{1.0} &  \cellcolor{Blu}0.745 &  \cellcolor{Blu}\textbf{0.972}  \\
        \hline
        \multirow{11}{*}{\emph{DDI}} & Gaifman & \cellcolor{LightBlue}0.469 & \cellcolor{LightBlue}0.707 & \cellcolor{LightBlue}0.564 & \cellcolor{LightBlue}0.581\\
        & Neural-LP$_3$ & \cellcolor{LightBlue}0.727 & \cellcolor{LightBlue}0.336 & \cellcolor{LightBlue}0.459 & \cellcolor{LightBlue}0.368\\
        & Neural-LP$_{10}$ & \cellcolor{LightBlue}0.779 & \cellcolor{LightBlue}0.338 & \cellcolor{LightBlue}0.472 & \cellcolor{LightBlue}0.403\\
        & metapath2vec & \cellcolor{LightBlue}0.782 & \cellcolor{LightBlue}0.652 & \cellcolor{LightBlue}0.711 & \cellcolor{LightBlue}0.707\\
        & PRAGCN & \cellcolor{LightBlue}0.427 & \cellcolor{LightBlue}0.700 & \cellcolor{LightBlue}0.531 & \cellcolor{LightBlue}0.695\\
        \cline{2-6}
        & ComplEx & \cellcolor{LightGreen}0.832 & \cellcolor{LightGreen}0.492 & \cellcolor{LightGreen}0.618 & \cellcolor{LightGreen}0.705\\
        & ConvE & \cellcolor{LightGreen}0.931 & \cellcolor{LightGreen}0.384 & \cellcolor{LightGreen}0.544 & \cellcolor{LightGreen}0.678\\
        & SimplE & \cellcolor{LightGreen}0.992 & \cellcolor{LightGreen}0.288 & \cellcolor{LightGreen}0.446 & \cellcolor{LightGreen}0.503\\
        & N+LF & \cellcolor{LightGreen}0.682 & \cellcolor{LightGreen}0.924 & \cellcolor{LightGreen}0.785 & \cellcolor{LightGreen}0.781\\
        \cline{2-6}
        & R-GCN & \cellcolor{LightPurple}0.571 & \cellcolor{LightPurple}\textbf{1.0} & \cellcolor{LightPurple}0.727 & \cellcolor{LightPurple}0.922 \\
        & CompGCN & \cellcolor{LightPurple}0.882 & \cellcolor{LightPurple}0.552 & \cellcolor{LightPurple}0.679 & \cellcolor{LightPurple}0.826\\
        \cline{2-6}
        & \textbf{RD$^2$GCN} &  \cellcolor{Blu}\textbf{0.998} &  \cellcolor{Blu}0.986 &  \cellcolor{Blu}\textbf{0.992} &  \cellcolor{Blu}\textbf{0.998}\\
        \hline
        \end{tabular}}
    \label{tab:results_link}
    \end{minipage}
\end{table}

\textbf{(Q1. Smaller data sets)} Table \ref{tab:results_link} shows the result of link prediction task. Our method outperforms all the baselines significantly in 2 of the 3 data sets with the difference being significant in the smaller data set \emph{ICML'18} and is comparable in the \emph{ICLR} data set. Note that although the recall is high for the neural embedding baselines, the corresponding F1 score and AUC-PR are low which implies that the {\bf baseline relational embedding methods have a high rate of false positives}. This clearly demonstrates that RD$^2$GCN is significantly better than the strong baselines for the link prediction task. This answers \textbf{Q1} affirmatively.

\textbf{(Q2. Secondary graph/distance matrix impact)} The main advantage of our method is learning a secondary graph structure where both link prediction and node classification tasks become simple prediction tasks in this new graph. As can be seen from the results for link prediction, a simple discriminative machine learning algorithm (logistic regression), used on top of the learned features (N+LF) performs better than the other baselines including GCN-based baselines. In case of node classification, the results are comparable. 

\begin{figure}[!htb]
    \centering
\begin{minipage}[b]{.47\linewidth}
    \centering    
  \includegraphics[width=\columnwidth]{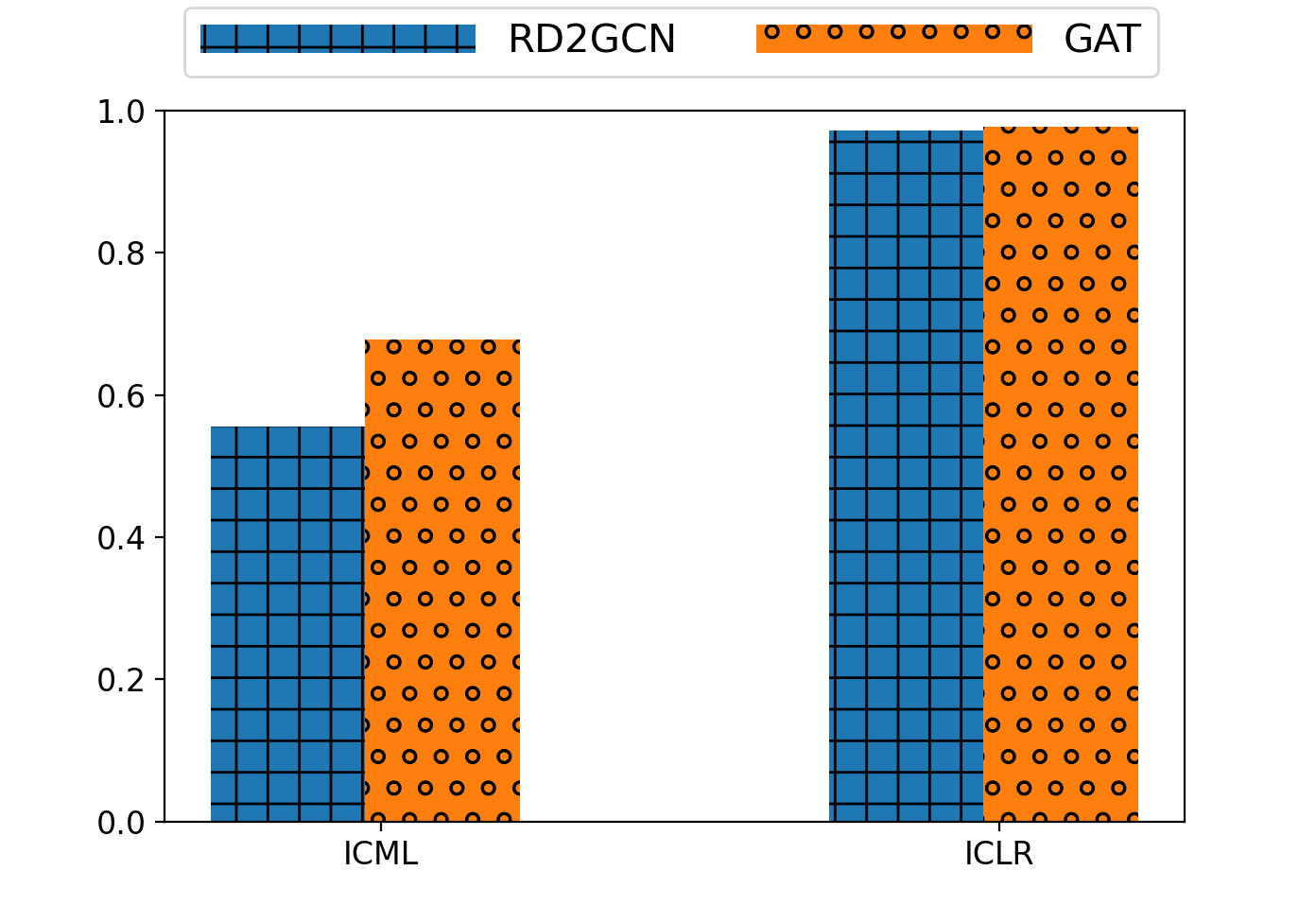}
  \caption{Comparison (AUC-PR) with GATs. 
  DDI does not run using GAT. 
  }
    \label{fig:gatvsgcn}
\end{minipage}
\hfill
\begin{minipage}[b]{.45\linewidth}
    \centering
    \includegraphics[width=\columnwidth]{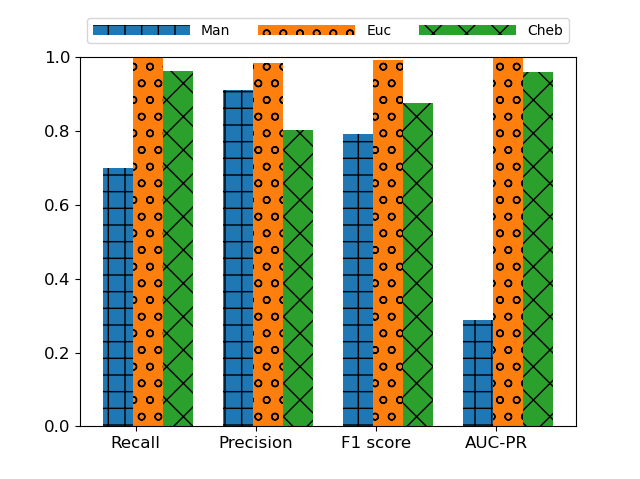}
    \caption{Effect of distance measures on the DDI data set.}
    \label{fig:dist}
\end{minipage}
\end{figure}

We also compare our method with Graph Attention Networks (GATs) \cite{velivckovic2017graph} which uses the our learned featues with the adjacency matrix instead of a distance matrix. Figure \ref{fig:gatvsgcn} shows the results and it can be seen that using a distance matrix is also effective. This is expected since in the secondary structure, as the nodes show the query, there is no particular notion of connection between the nodes. This answers \textbf{Q2} affirmatively, learning a secondary graph structure is useful for prediction tasks.

\textbf{(Q3. SRL + GCN)} Our results show that using SRL models (relational density estimation in our case) as the underlying feature learner which are then fed to a neural model, GCN, gives us a powerful hybrid model that can be used seamlessly with relational data. Using a SRL model as the initial layer of a neural model results in learning richer initial features set used by the neural model. This initial feature set can take advantage of underlying graph structure faithfully and thus, in accordance, {\bf leads to the neural model learning far richer abstract features which in turn leads to better predictive performance}. Our evaluations on both tasks support this as our method significantly outperforms GCN baselines in all domains, answering \textbf{Q3} affirmatively. Neural SRL (neuro-symbolic) models can be both interpretable and expressive.

\textbf{(Q4. Effective rule learning)} To answer \textbf{Q4}, we use 4 different rule learning methods: handwritten rules (Gaifman), NeuralLP (rule length 3 \& 10), metapath2vec and PRA. Table \ref{tab:results_link} clearly demonstrate that using our {\em density estimation method significantly outperforms all rule learning method across all domains}. Comparing PRAGCN and RD$^2$GCN is especially interesting since this shows that rule learning method plays a crucial role in learning richer features, especially in the imbalanced domains, where relational density estimation is demonstrably beneficial since both methods share the underlying GCN. The difference in performance of PRAGCN and RD$^2$GCN is {\bf significantly high} in the highly imbalanced domains \emph{ICML'18} and \emph{ICLR} where the features learned by the PRA method result in all examples being classified as negative.

\textbf{(Q5. Effect of distance measures)} Figure \ref{fig:dist} presents the effect of 2 other distance measures, Manhattan ($L_1$) and Chebyshev ($L_\infty$), in addition to Euclidean ($L_2$) on the performance of RD$^2$GCN on the DDI data set. Since Euclidean is the straight line i.e. shortest distance between nodes, it performs the best as expected. This answers \textbf{Q5}.

\textbf{(Q6. Effect of parameter choices)} To answer \textbf{Q6}, we change the size of the hidden layers in the GCN as well as the number of hidden layers and test our method on \emph{ICML'18} and \emph{DDI} data sets. Tables \ref{tab:hid_size} and \ref{tab:hid_layers} show that change of these parameters have none or very minuscule effect on the overall results. This answers \textbf{Q6} and also shows that the \emph{\textbf{learned features by themselves are quite expressive thus removing the need for a more complex GCN.}}

\begin{table}
\begin{minipage}[t]{\columnwidth}
    \caption{Effect of change in size of hidden layers.}
     \scalebox{0.9}{
    \begin{tabular}{|c|c|c|c|c|c|}
        \hline
         Data & Size & Recall & Precision & F1 Score & AUC-PR\\
        \hline
        \multirow{3}{*}{\emph{ICML'18}} & 32 & 0.369 & 1.0 & 0.539 & 0.692\\
        &64 & 0.369 & 1.0 & 0.539 & 0.692\\
        &128 & 0.369 & 1.0 & 0.539 & 0.692\\
        \hline
        \multirow{3}{*}{\emph{DDI}} & 32 & 0.989 & 0.998 & 0.993 & 0.998\\
        &64 & 0.989 & 0.998 & 0.993 & 0.998\\
        &128 & 0.989 & 0.998 & 0.993 & 0.998\\
        \hline
        \end{tabular}}
    \label{tab:hid_size}
\end{minipage}
\hspace{0.4cm}
\begin{minipage}[t]{\columnwidth}
    \caption{Effect of change in \# of hidden layers.}
    \scalebox{0.9}{
    \begin{tabular}{|c|c|c|c|c|c|}
        \hline
         Data & \# & Recall & Precision & F1 Score & AUC-PR\\
        \hline
        \multirow{3}{*}{\emph{ICML'18}} & 3 & 0.369 & 1.0 & 0.539 & 0.692\\
        & 4 & 0.369 & 1.0 & 0.539 & 0.692\\
        & 5 & 0.369 & 1.0 & 0.539 & 0.692\\
        \hline
        \multirow{3}{*}{\emph{DDI}} & 3 & 0.989 & 0.998 & 0.994 & 0.999\\
        & 4 & 1.0 & 0.997 & 0.998 & 0.999\\
        & 5 & 0.999 & 0.991 & 0.995 & 0.997\\
        \hline
        \end{tabular}}
    \label{tab:hid_layers}
    \end{minipage}
 \end{table}

\section{Conclusion}
We presented the first GCN method that can learn from multi-relational data utilizing the different densities separately. Our method does not make assumptions on the supervision or the arity of predicates and automatically constructs rules that allow for a rich latent representation. We significantly outperform the recently successful methods on the link prediction task across multiple data sets. Allowing for joint learning and inference over multiple types of relations is an important future direction. Using more classical rule learning techniques such as \cite{quinlan1990learning,muggleton1995inverse,srinivasan2001aleph} is another interesting direction. Extending our framework to node classification is the natural next step. Finally, learning in the presence of hidden/latent data and rich human domain knowledge is essential for deploying SRL methods in real tasks.

\bibliography{aaai22}

\begin{thebibliography}{47}
\providecommand{\natexlab}[1]{#1}

\bibitem[{Bai et~al.(2019)Bai, Ding, Bian, Chen, Sun, and Wang}]{bai2019simgnn}
Bai, Y.; Ding, H.; Bian, S.; Chen, T.; Sun, Y.; and Wang, W. 2019.
\newblock Simgnn: A neural network approach to fast graph similarity
  computation.
\newblock In \emph{WSDM}.

\bibitem[{Bai et~al.(2018)Bai, Ding, Sun, and Wang}]{bai2018convolutional}
Bai, Y.; Ding, H.; Sun, Y.; and Wang, W. 2018.
\newblock Convolutional set matching for graph similarity.
\newblock \emph{NIPS}.

\bibitem[{Balazevic, Allen, and Hospedales(2019)}]{balazevic2019multi}
Balazevic, I.; Allen, C.; and Hospedales, T. 2019.
\newblock Multi-relational Poincar{\'e} graph embeddings.
\newblock In \emph{NIPS}.

\bibitem[{Bala{\v{z}}evi{\'c}, Allen, and
  Hospedales(2019{\natexlab{a}})}]{balavzevic2019hypernetwork}
Bala{\v{z}}evi{\'c}, I.; Allen, C.; and Hospedales, T.~M. 2019{\natexlab{a}}.
\newblock Hypernetwork knowledge graph embeddings.
\newblock In \emph{ICANN}.

\bibitem[{Bala{\v{z}}evi{\'c}, Allen, and
  Hospedales(2019{\natexlab{b}})}]{balavzevic2019tucker}
Bala{\v{z}}evi{\'c}, I.; Allen, C.; and Hospedales, T.~M. 2019{\natexlab{b}}.
\newblock Tucker: Tensor factorization for knowledge graph completion.
\newblock \emph{EMNLP}.

\bibitem[{Bordes et~al.(2013)Bordes, Usunier, Garcia-Duran, Weston, and
  Yakhnenko}]{bordes2013translating}
Bordes, A.; Usunier, N.; Garcia-Duran, A.; Weston, J.; and Yakhnenko, O. 2013.
\newblock Translating embeddings for modeling multi-relational data.
\newblock In \emph{NIPS}.

\bibitem[{Cai, Zheng, and Chang(2018)}]{cai2018comprehensive}
Cai, H.; Zheng, V.~W.; and Chang, K. C.-C. 2018.
\newblock A comprehensive survey of graph embedding: Problems, techniques, and
  applications.
\newblock \emph{TKDE}.

\bibitem[{Cohen(2016)}]{cohen2016tensorlog}
Cohen, W.~W. 2016.
\newblock Tensorlog: A differentiable deductive database.
\newblock \emph{arXiv preprint arXiv:1605.06523}.

\bibitem[{Cvetkovic et~al.(1980)Cvetkovic, Doob, Sachs
  et~al.}]{cvetkovic1980spectra}
Cvetkovic, D.~M.; Doob, M.; Sachs, H.; et~al. 1980.
\newblock \emph{Spectra of graphs}.

\bibitem[{Defferrard, Bresson, and
  Vandergheynst(2016)}]{defferrard2016convolutional}
Defferrard, M.; Bresson, X.; and Vandergheynst, P. 2016.
\newblock Convolutional neural networks on graphs with fast localized spectral
  filtering.
\newblock In \emph{NIPS}.

\bibitem[{Dettmers et~al.(2018)Dettmers, Minervini, Stenetorp, and
  Riedel}]{dettmers2018convolutional}
Dettmers, T.; Minervini, P.; Stenetorp, P.; and Riedel, S. 2018.
\newblock Convolutional 2d knowledge graph embeddings.
\newblock In \emph{AAAI}.

\bibitem[{Dhami et~al.(2018)Dhami, Kunapuli, Das, Page, and
  Natarajan}]{dhami2018drug}
Dhami, D.~S.; Kunapuli, G.; Das, M.; Page, D.; and Natarajan, S. 2018.
\newblock Drug-drug interaction discovery: kernel learning from heterogeneous
  similarities.
\newblock \emph{Smart Health}.

\bibitem[{Dong, Chawla, and Swami(2017)}]{dong2017metapath2vec}
Dong, Y.; Chawla, N.~V.; and Swami, A. 2017.
\newblock metapath2vec: Scalable representation learning for heterogeneous
  networks.
\newblock In \emph{KDD}.

\bibitem[{Gaifman(1982)}]{gaifman1982local}
Gaifman, H. 1982.
\newblock On local and non-local properties.
\newblock In \emph{Studies in Logic and the Foundations of Mathematics}.

\bibitem[{Getoor and Taskar(2007)}]{getoor2007introduction}
Getoor, L.; and Taskar, B. 2007.
\newblock \emph{Introduction to statistical relational learning}.
\newblock MIT press.

\bibitem[{He et~al.(2015)He, Liu, Ji, and Zhao}]{he2015learning}
He, S.; Liu, K.; Ji, G.; and Zhao, J. 2015.
\newblock Learning to represent knowledge graphs with gaussian embedding.
\newblock In \emph{CIKM}.

\bibitem[{Jaeger(2007)}]{jaeger2007RBNs}
Jaeger, M. 2007.
\newblock Parameter learning for relational bayesian networks.
\newblock In \emph{ICML}.

\bibitem[{Kaur et~al.(2019)Kaur, Kunapuli, Joshi, Kersting, and
  Natarajan}]{kaur2019neural}
Kaur, N.; Kunapuli, G.; Joshi, S.; Kersting, K.; and Natarajan, S. 2019.
\newblock Neural networks for relational data.
\newblock In \emph{ILP}.

\bibitem[{Kazemi and Poole(2018)}]{kazemi2018simple}
Kazemi, S.~M.; and Poole, D. 2018.
\newblock Simple embedding for link prediction in knowledge graphs.
\newblock In \emph{NeurIPS}.

\bibitem[{Khot, Natarajan, and Shavlik(2014)}]{khot2014relational}
Khot, T.; Natarajan, S.; and Shavlik, J. 2014.
\newblock Relational one-class classification: A non-parametric approach.
\newblock In \emph{AAAI}.

\bibitem[{Kipf and Welling(2017)}]{kipf2017semi}
Kipf, T.~N.; and Welling, M. 2017.
\newblock Semi-supervised classification with graph convolutional networks.
\newblock \emph{ICLR}.

\bibitem[{Kolyvakis, Kalousis, and Kiritsis(2019)}]{kolyvakis2019hyperkg}
Kolyvakis, P.; Kalousis, A.; and Kiritsis, D. 2019.
\newblock HyperKG: Hyperbolic Knowledge Graph Embeddings for Knowledge Base
  Completion.
\newblock \emph{arXiv preprint arXiv:1908.04895}.

\bibitem[{Lao and Cohen(2010)}]{lao2010relational}
Lao, N.; and Cohen, W.~W. 2010.
\newblock Relational retrieval using a combination of path-constrained random
  walks.
\newblock \emph{Machine learning}.

\bibitem[{Li et~al.(2019)Li, Gu, Dullien, Vinyals, and Kohli}]{li2019graph}
Li, Y.; Gu, C.; Dullien, T.; Vinyals, O.; and Kohli, P. 2019.
\newblock Graph matching networks for learning the similarity of graph
  structured objects.
\newblock \emph{ICML}.

\bibitem[{Muggleton(1995)}]{muggleton1995inverse}
Muggleton, S. 1995.
\newblock Inverse entailment and Progol.
\newblock \emph{New generation computing}.

\bibitem[{Natarajan et~al.(2008)Natarajan, Tadepalli, Dietterich, and
  Fern}]{natarajan2008Combining}
Natarajan, S.; Tadepalli, P.; Dietterich, T.~G.; and Fern, A. 2008.
\newblock Learning first-order probabilistic models with combining rules.
\newblock \emph{Annals of Mathematics and Artificial Intelligence}.

\bibitem[{Nickel, Rosasco, and Poggio(2016)}]{nickel2016holographic}
Nickel, M.; Rosasco, L.; and Poggio, T. 2016.
\newblock Holographic embeddings of knowledge graphs.
\newblock In \emph{AAAI}.

\bibitem[{Nickel, Tresp, and Kriegel(2011)}]{nickel2011three}
Nickel, M.; Tresp, V.; and Kriegel, H.-P. 2011.
\newblock A three-way model for collective learning on multi-relational data.
\newblock In \emph{ICML}.

\bibitem[{Niepert(2016)}]{niepert2016discriminative}
Niepert, M. 2016.
\newblock Discriminative gaifman models.
\newblock In \emph{NIPS}.

\bibitem[{Quinlan(1990)}]{quinlan1990learning}
Quinlan, J.~R. 1990.
\newblock Learning logical definitions from relations.
\newblock \emph{Machine learning}.

\bibitem[{Raedt et~al.(2016)Raedt, Kersting, Natarajan, and
  Poole}]{raedt2016statistical}
Raedt, L.~D.; Kersting, K.; Natarajan, S.; and Poole, D. 2016.
\newblock Statistical relational artificial intelligence: Logic, probability,
  and computation.
\newblock \emph{Synthesis Lectures on Artificial Intelligence and Machine
  Learning}.

\bibitem[{Richardson and Domingos(2006)}]{richardson2006markov}
Richardson, M.; and Domingos, P. 2006.
\newblock Markov logic networks.
\newblock \emph{Machine learning}.

\bibitem[{Rouvray and Balaban(1979)}]{rouvray1979chemical}
Rouvray, D.~H.; and Balaban, A.~T. 1979.
\newblock Chemical applications of graph theory.
\newblock \emph{Applications of Graph Theory}.

\bibitem[{Schlichtkrull et~al.(2018)Schlichtkrull, Kipf, Bloem, Van Den~Berg,
  Titov, and Welling}]{schlichtkrull2018modeling}
Schlichtkrull, M.; Kipf, T.~N.; Bloem, P.; Van Den~Berg, R.; Titov, I.; and
  Welling, M. 2018.
\newblock Modeling relational data with graph convolutional networks.
\newblock In \emph{European Semantic Web Conference}.

\bibitem[{Sinha et~al.(2015)Sinha, Shen, Song, Ma, Eide, Hsu, and
  Wang}]{sinha2015overview}
Sinha, A.; Shen, Z.; Song, Y.; Ma, H.; Eide, D.; Hsu, B.-J.; and Wang, K. 2015.
\newblock An overview of microsoft academic service (mas) and applications.
\newblock In \emph{WWW}.

\bibitem[{Srinivasan(2001)}]{srinivasan2001aleph}
Srinivasan, A. 2001.
\newblock The aleph manual.

\bibitem[{Sun et~al.(2019)Sun, Deng, Nie, and Tang}]{sun2019rotate}
Sun, Z.; Deng, Z.-H.; Nie, J.-Y.; and Tang, J. 2019.
\newblock Rotate: Knowledge graph embedding by relational rotation in complex
  space.
\newblock \emph{ICLR}.

\bibitem[{Toutanova and Chen(2015)}]{toutanova2015observed}
Toutanova, K.; and Chen, D. 2015.
\newblock Observed versus latent features for knowledge base and text
  inference.
\newblock In \emph{CVSC Workshop at ACL}.

\bibitem[{Trouillon et~al.(2016)Trouillon, Welbl, Riedel, Gaussier, and
  Bouchard}]{trouillon2016complex}
Trouillon, T.; Welbl, J.; Riedel, S.; Gaussier, {\'E}.; and Bouchard, G. 2016.
\newblock Complex embeddings for simple link prediction.
\newblock ICML.

\bibitem[{Vashishth et~al.(2020)Vashishth, Sanyal, Nitin, and
  Talukdar}]{vashishth2020composition}
Vashishth, S.; Sanyal, S.; Nitin, V.; and Talukdar, P. 2020.
\newblock Composition-based multi-relational graph convolutional networks.
\newblock \emph{ICLR}.

\bibitem[{Veli{\v{c}}kovi{\'c} et~al.(2018)Veli{\v{c}}kovi{\'c}, Cucurull,
  Casanova, Romero, Lio, and Bengio}]{velivckovic2017graph}
Veli{\v{c}}kovi{\'c}, P.; Cucurull, G.; Casanova, A.; Romero, A.; Lio, P.; and
  Bengio, Y. 2018.
\newblock Graph attention networks.
\newblock \emph{ICLR}.

\bibitem[{Wang et~al.(2017)Wang, Mao, Wang, and Guo}]{wang2017knowledge}
Wang, Q.; Mao, Z.; Wang, B.; and Guo, L. 2017.
\newblock Knowledge graph embedding: A survey of approaches and applications.
\newblock \emph{IEEE Transactions on Knowledge and Data Engineering}.

\bibitem[{Wang et~al.(2014)Wang, Zhang, Feng, and Chen}]{wang2014knowledge}
Wang, Z.; Zhang, J.; Feng, J.; and Chen, Z. 2014.
\newblock Knowledge graph embedding by translating on hyperplanes.
\newblock In \emph{AAAI}.

\bibitem[{Xiao, Huang, and Zhu(2016)}]{xiao2016transg}
Xiao, H.; Huang, M.; and Zhu, X. 2016.
\newblock TransG: A generative model for knowledge graph embedding.
\newblock In \emph{ACL}.

\bibitem[{Yang et~al.(2015)Yang, Yih, He, Gao, and Deng}]{yang2015embedding}
Yang, B.; Yih, W.-t.; He, X.; Gao, J.; and Deng, L. 2015.
\newblock Embedding entities and relations for learning and inference in
  knowledge bases.
\newblock \emph{ICLR}.

\bibitem[{Yang, Yang, and Cohen(2017)}]{yang2017differentiable}
Yang, F.; Yang, Z.; and Cohen, W.~W. 2017.
\newblock Differentiable learning of logical rules for knowledge base
  reasoning.
\newblock In \emph{NIPS}.

\bibitem[{Zhang et~al.(2019)Zhang, Tay, Yao, and Liu}]{zhang2019quaternion}
Zhang, S.; Tay, Y.; Yao, L.; and Liu, Q. 2019.
\newblock Quaternion knowledge graph embeddings.
\newblock In \emph{NIPS}.

\end{thebibliography}

\end{document}